\documentclass[11pt]{amsart}
\usepackage{amsmath, amssymb,mathabx,mathrsfs}
\usepackage{amssymb}     \usepackage{graphicx}   
     \usepackage{amsmath}
\usepackage{color}
\usepackage{tcolorbox}
\usepackage{hyperref}
\usepackage[toc,page]{appendix}
\usepackage{stackengine}
\usepackage{comment}
\usepackage[multiple]{footmisc}
\usepackage{bigfoot}
\usepackage{float}
\usepackage{epsfig}
\usepackage{graphicx}
\usepackage[T1]{fontenc}
\usepackage[utf8]{inputenc}
\usepackage[english]{babel}
\usepackage{amsmath}
\usepackage{graphicx}
\usepackage[Export]{adjustbox}
\usepackage{longtable}

\usepackage{subcaption}
\usepackage{mathtools, nccmath}
\usepackage{amsthm}
\usepackage{multicol}
\usepackage{tikz,pgfplots}
\usepackage[normalem]{ulem}

\theoremstyle{plain}
\newtheorem{thm}{Theorem}[section]

\newtheorem{defn}[thm]{Definition}
\theoremstyle{remark}

\newtheorem{ex}[thm]{Example}

\theoremstyle{definition}
\theoremstyle{remark}
\newtheorem{remark}{Remark}[section]

\adjustboxset{width=0.3\linewidth}
\renewcommand{\epsilon}{\varepsilon}

\def\R{{\mathbb R}}

\DeclareNewFootnote{AAffil}[arabic]
\DeclareNewFootnote{ANote}[fnsymbol]
\pgfplotsset{compat=1.17} 
\begin{document}
\title[Application of RMT to deep leaning]{Deep Learning Weight Pruning with RMT-SVD: Increasing Accuracy and Reducing Overfitting}

\author{Yitzchak Shmalo, Jonathan Jenkins, Oleksii Krupchytskyi} 
\footnote{Department of Mathematics, Penn State University, USA. E-mail address: yms5281@psu.edu, jtj5311@psu.edu, omk5165@psu.edu. \newline

 Code used in this paper is available at \url{https://github.com/jtj5311/NN-RMT-SVD}}

 \begin{abstract}
   In this work, we present some applications of random matrix theory for the training of deep neural networks. Recently, random matrix theory (RMT) has been applied to the overfitting problem in deep learning. Specifically, it has been shown that the spectrum of the weight layers of a deep neural network (DNN) can be studied and understood using techniques from RMT.
  In this work, these RMT techniques will be used to determine which and how many singular values should be removed from the weight layers of a DNN during training, via singular value decomposition (SVD), so as to reduce overfitting and increase accuracy. We show the results on a simple DNN model trained on MNIST. In general, these techniques may be applied to any fully connected layer of a pretrained DNN to reduce the number of parameters in the layer while preserving and sometimes increasing the accuracy of the DNN.          
 
 

 \end{abstract}
       \maketitle

\section{Introduction}\label{into}

DNNs are a powerful tool in the classification problem, where they determine the class to which a set of objects $S \subset \R^n$ belongs. In this process, a training set $T \subset \R^n$ with known class labels is used to train the DNN using a loss function (for example the cross-entropy loss function \eqref{loss_function}), with the goal of improving accuracy as loss decreases. Accuracy refers to the percentage of correct classifications made by the DNN for elements in the training or test set. DNNs have been demonstrated to be effective in solving a wide range of real-world classification problems, including handwriting recognition \cite{LBD}, image classification \cite{krizhevsky2017imagenet}, speech recognition \cite{hinton2012deep}, and natural language processing \cite{sutskever2014sequence}.

However, a issue that arises when training DNNs is overfitting. Here we say that a model is overfitting  when, as training progresses, it becomes more accurate on the training set but less accurate on the test set or when the model becomes more accurate on the training set and plateaus on the test set. In the latter case, the model reaches a point where it can no longer improve its performance on the test set despite further training on the training set. This indicates that the model has learned to fit the training data too well, resulting in poor generalization performance.

Overfitting occurs when the model becomes too complex and starts memorizing the training data instead of generalizing to new data. This can lead to poor performance on the test set, even though the training set accuracy is high. To mitigate overfitting, various regularization techniques have been developed, such as dropout \cite{srivastava2014dropout} and early stopping \cite{prechelt2012early}.

In recent years, there has been a growing interest in using RMT for overfitting in deep learning \cite{martin2021implicit}. In this work, we present some applications of RMT for the training of DNNs. Some numerical experiments will be shown on simple DNN models trained on MNIST. In general, these RMT techniques can be applied to other types of DNNs and any fully connected layer of a pretrained DNN to reduce the number of parameters in the layer while preserving and sometimes increasing the accuracy of the DNN.

In previous work,  such as \cite{xu2019trained, yang2020learning,xue2013restructuring,cai2014fast,anhao2016svd}, thresholding has been used as a method to remove the small singular values of a weight matrix in a DNN to avoid overfitting. They found that removing small singular values in weight matrices, through SVD, could lead to improved performance and reduced overfitting. However, they did not use the Marchenko-Pastur (MP) distribution to determine a threshold, which is the  method used in this work. Instead, they used other methods such as energy ratio threshold and monitoring the test error to determine the optimal threshold for removing the singular values. We, therefore, study the use of the MP distribution to find a threshold in deep learning for pruning singular values during training. A threshold based on MP theory might provide insight into why previous singular values pruning techniques work as well as improve on pre-existing techniques. Using a similar threshold method, \cite{staats2022boundary} showed how one can improve the accuracy of DNNs trained on noisy data.

The rest of the paper is organized as follows. In Section 2, we will delve deeper into the problem of overfitting in DNNs and present some of the commonly used regularization techniques to address this issue. 

In Section 3, we will provide a more comprehensive overview of RMT, including some key concepts. 

In Section 4, we will present the results of our experiments, where we applied RMT techniques to DNN models trained on MNIST. These results will demonstrate the effectiveness of using RMT for overfitting in deep learning and the potential for improving the accuracy of DNNs. It is important to note that in out numerical experiments we deliberately start with DNNs which are overparameterized with respect to the data set MNIST and don't preform well when trained on MNIST. The goal is to see whether our techniques can be used to reduce the parameters of the DNN, throughout training, so that it stops overfitting.    

Finally, in Section 5, we will provide a summary of our findings and discuss the future directions for this research. This may include exploring new ways to apply RMT to deep learning and investigating the scalability of these techniques to larger datasets and models. It would also be useful to study how RMT can improve on SVD based DNN regularization, such as nuclear regularization found in \cite{xu2019trained},   to  provide even better performance.

This application of RMT to deep learning is an exciting area of research with many potential benefits. In this work, we hope to shed some light on the potential of this approach and demonstrate its effectiveness, at least on a small data set with simple DNNs, in improving the accuracy and reducing overfitting in deep neural networks.

	 \section{Preliminaries}

DNNs are a popular tool for solving the classification problem, where a set of objects $S \subset \R^n$ is assigned to one of $K$ classes. The goal is to approximate an exact classifier $\phi^*$ which maps $s \in T \subset S$ to a vector of probabilities $(p_1(s), \dots, p_K(s))$, where $p_{i(s)}=1$ and $p_{j}=0$ for $j \neq i(s)$, and $i(s)$ denotes the correct class of $s$. The exact classifier $\phi^*$ is only known for a training set $T$, and so DNNs are trained to approximate $\phi^*$ by constructing a parameterized classifier $\phi(\alpha, s)$ with the aim of extending $\phi^*$ from $T$ to all of $S$ via $\phi(\alpha, s)$.

This is achieved by finding parameters $\alpha$ such that $\phi(\alpha, s)$ maps $s \in T$ to the same class as $\phi^*$ while still allowing the classifier to generalize to elements of $s \in S$. The parameters $\alpha$ are optimized minimizing a loss function, with the goal of improving accuracy as the loss decreases.

 In this work, a DNN is represented as a composition of two functions: the softmax function $\rho$ and an intermediate function $X(\cdot,\alpha)$. The function $X(\cdot,\alpha)$ is defined as a composition of affine transformations and nonlinear activations, as follows:
\begin{itemize}
\item $M_l(\cdot,\alpha_l)$ is an affine function that maps $\mathbb R^{N_{l-1}}$ to $\mathbb R^{N_l}$, and depends on a parameter matrix $W_l$ of size $N_{l-1} \times N_l$ and a bias vector $\beta_l$ (i.e. $M_l(x)=W_l(x)+\beta_l$).
\item $\lambda: \R^m \mapsto \R^m$ is a nonlinear activation function. In this paper, we assume that $\lambda$ is the ReLU activation function applied to every coordinate.
\item $X(\cdot,\alpha)=\lambda \circ M_k\cdots \lambda \circ M_1$, where $k$ is the number of layers in the DNN. Note, each $\lambda$ here might be different from the other given that the domains of each is different. 
\end{itemize}

Finally, $\rho$ is the softmax function, which normalizes the output of $X(\cdot,\alpha)$ into probabilities. The components of $\rho$ are calculated as:
\begin{equation}
\label{soft_max}
\rho_i(s,\alpha)=\frac{\exp(X_i(s,\alpha))}{\sum_{i=1}^{K}\exp(X_i(s,\alpha))}.
\end{equation}

The output of the DNN $\phi$ is a vector representing the probabilities of an object $s \in T$ belonging to a certain class $i$. $\phi=\phi(s,\alpha)$, where $\alpha \in \R^\nu$ is the parameter space of the DNN and $\nu \gg 1$ is the dimension of the parameter space. The goal is to train the DNN $\phi$ to approximate the exact classifier by minimizing a loss function, such as the cross-entropy loss function
\begin{equation}
\bar L(\alpha)=-\frac{1}{|T|}\sum_{s\in T} \log\left(p_{i(s)}(s,\alpha)\right).
\label{loss_function}
\end{equation}

However, even if the loss function is minimized, it is still possible that the DNN $\phi$ will not generalize well to new data points $s \notin T$. This is due to the fact that DNNs have a large capacity to fit the training data and can sometimes fit the noise in the data, leading to overfitting as will be explained in more detail in Subsection \ref{overfitting}. To combat overfitting, several approaches have been developed, including regularization techniques, such as dropout, early stopping, and weight decay, as well as architectural choices, such as using smaller networks, or using different activation functions.

	 \begin{subsection} 
  {What is overfitting and why it is bad?}\label{overfitting}

   One of the mathematical concepts behind the generalization issue in DNNs is the notion of overfitting. Overfitting occurs when the model fits the training data too closely and ends up memorizing the training data instead of learning the underlying patterns in the data. This leads to poor performance on unseen data as the model has not learned to generalize from the training data to unseen data. 

   To identify if overfitting is occurring, it is common to monitor the performance of the model on both the training and test datasets during the training process. If the model's performance on the training data continues to improve but the performance on the test data starts to plateau or even worsen, it is a sign that the model is overfitting.

In such cases, the model has become too specialized in recognizing the features and patterns present in the training data, to the point where it cannot generalize to new data. The model has essentially "memorized" the training data, and is unable to extract more generalized knowledge that can be applied to new, unseen data.

	   A simple example of overfitting and underfitting can be found in Fig. \ref{over_under_fitting}. The goal is to find the separation line between the blue and yellow dots. The figure shows three cases.  
		      In the underfitting case only two parameters define the dashed line. In the overfitting case, too many parameters define the dashed line. This means that a new data point from the test set is likely to be misclasified. In this example, a classifier might have high accuracy on training set and low accuracy on test set. This example also illustrates how a DNN with too many parameters might overfit. This is because the 'extra'  parameters might 'learn' from the randomness in the data, thus learning structure which is not there.

    	\begin{figure}
		\includegraphics[width=1.1\textwidth]{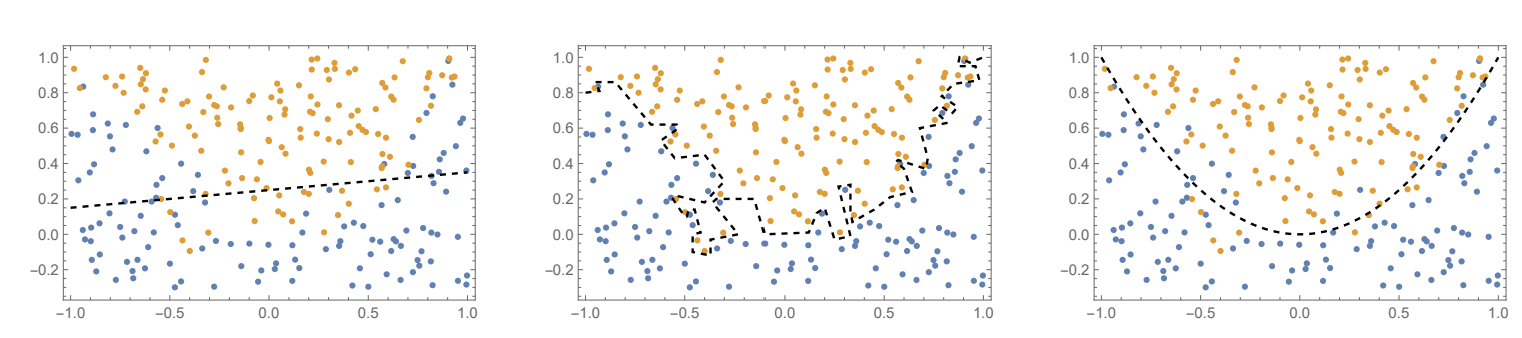}
    	\caption{Left figure shows underfitting: too few parameters (e.g., 2 parameters for straight line) making many mistakes on the classification.
		    Middle figure shows overfitting: too many parameters lead to curve perfectly separating orange data from blue, but we are likely to misclassify new data points. Separating curve captures too many details of the data when describing the underlying structure of the data.
		    Right figure shows good fit: not too many parameters are used and not too few. The classification makes some mistakes, but clearly captures the data's underlying structure.}
    	\label{over_under_fitting}
    	\end{figure}

Thus, this example highlights how a DNN with an excessive number of parameters can easily overfit due to the extra parameters that might capture random noise in the data instead of real patterns. Such parameters would be random, meaning that the overfitting in DNNs might be related to the randomness of parameters in the weight layers $M_l$ of the DNN. 
	There are at least two sources of randomness in a DNN weight layer matrix after training. Before training starts, DNN parameters $\alpha(0)$ are chosen randomly. Then when applying gradient decent  (GD) we have an update to the parameters given by the following equation:
	\begin{equation}
		\alpha(n+1)=\alpha(n)-\tau\nabla L(\alpha(n)),
	\end{equation}

 where $\tau$ is the step size. The loss gradient $\nabla L(\alpha(n))$ is determined by the training data $T$ and so is mostly deterministic, given that the data $T$ is assumed to be deterministic. Thus, the random DNN parameters are gradually replaced with deterministic parameters. However, the training set $T$ is only \emph{mostly} deterministic. Each object $s\in T$ is sampled from a random variable, so $T$ contains some randomness. Further, because the parameters $\alpha$ started out being random even after some training they might still have some random structure to them. In practice, the randomness of $\alpha(n)$ decreases as $n\to\infty$, \cite{martin2021implicit}, but some randomness remains. As mentioned, DNN parameters are arranged in \emph{weight matrices} and bais vectors. Thus, it has been shown that one can use the well-developed theory of random matrices to study the randomness of the parameters $\alpha$ and characterize and avoid overfitting, see \cite{martin2021implicit,martin2021predicting,staats2022boundary}.

 The capacity of a DNN model is also an important concept when studying overfitting. The capacity of a DNN refers to its ability to fit a wide range of functions. A model with a high capacity has a large number of parameters and can fit complex functions, while a model with low capacity has limited parameters and can only fit simpler functions. The generalization ability of a model is closely related to its capacity. If the model has too high a capacity, it can easily overfit the training data, while if the capacity is too low, it will not be able to fit the training data well. The capacity of a DNN is also related to the double decent phenomena, discussed in Subsection \ref{double_decent}. 

A method to control the capacity of a DNN is through regularization. Regularization adds constraints to the model that prevent it from overfitting the training data. One of the most common forms of regularization for DNNs is $L1$ and $L2$ regularization, which add penalty terms to the loss function that encourage the parameters to be small. Another form of regularization is dropout, which randomly drops out neurons during training to prevent overfitting.

Finally, the optimization algorithm used to train the model is also relevant to the generalization problem. The optimization algorithm updates the parameters of the model in order to minimize the loss function. Common optimization algorithms include gradient descent, stochastic gradient descent, and ADAM and use the gradient of the loss function with respect to the parameters to update the parameters. These algorithms have been proven to be effective for training DNNs, however, the choice of optimization algorithm or hyperparameters of the optimization algorithm, such as batch size or step size, can impact the generalization ability of the model, see \cite{martin2021implicit}.

In conclusion, when one overparameterizes a  DNN there is a good chance some of the extra parameters will retain some random characteristics either because they trained on the noise in the training set or because they were initially random. Thus, it might be possible to reduce overfitting in DNNs by removing some of these parameters. This would also allow us to increase accuracy of DNNs by removing  parameters so continuing training is easier given that the DNN is smaller. For more on the overparameterization of DNNs, see \cite{ma2018power,kawaguchi2017generalization,cohen2021learning}.

\subsection{Double descent phenomenon}
\label{double_decent}

Double descent is a phenomenon in DNNs that refers to the unexpected behavior of the test error, which initially decreases as model complexity increases, then increases again before finally decreasing to zero. The first descent occurs in the underparameterized classical regime when the model starts to learn the underlying patterns in the training data, while the second descent occurs in the overparameterized modern interpolation regime when the model has a large number of parameters. This behavior is counterintuitive because conventional wisdom suggests that increasing the complexity of the model would lead to overfitting and higher test error.

The double descent phenomenon has been observed in a variety of deep learning models, including convolutional neural networks (CNNs), residual networks (ResNets), and recurrent neural networks (RNNs), among others \cite{belkin2019reconciling,nakkiran2021deep,advani2020high}. It has also been shown to occur across a range of datasets and architectures, indicating that it is a fundamental property of DNNs.

When the parameters of a DNN grow drastically, the double descent phenomenon becomes more pronounced \cite{advani2020high}. In this regime, the test error first decreases as the model complexity increases due to the model's ability to capture more complex patterns in the training data. However, as the number of parameters continues to increase, the test error starts to rise due to overfitting. When the number of parameters is very large, the test error will drop again. One explanation is that DNNs which have many parameters can learn simple functions as well as complex functions, and may even prefer to learn simple functions over more complex ones \cite{belkin2019reconciling}. 

The double descent phenomenon has significant implications for the design and interpretation of DNNs. It suggests that increasing the complexity of the model beyond a certain threshold can lead to a increase in performance. Moreover, the double descent phenomenon challenges the conventional wisdom that regularization is always necessary to prevent overfitting. Instead, it suggests that models can benefit from having a small amount of overfitting, as it can help the model learn the underlying patterns in the data. 

Overall, the double descent phenomenon highlights the importance of understanding the fundamental properties of DNNs to design models that can generalize well and avoid overfitting. As the number of parameters in DNNs continues to grow, it is important to explore new regularization techniques and model architectures that can harness the power of large models while avoiding overfitting \cite{belkin2019reconciling}.

Because of this it is important to find techniques for preventing overfitting and improving the generalization performance of DNNs. The most common regularization techniques include weight decay, dropout, and early stopping. Weight decay and dropout introduce a penalty on the complexity of the model, while early stopping stops the training process before the model starts to overfit. These techniques have been shown to be effective in practice, but they may not be optimal for all problems.

One alternative approach to regularization is to use SVD as a  way of lowering the capacity of the DNN. SVD can be used to decompose the weight matrix of a DNN into a product of three matrices, which can be used to approximate the original weight matrix with a lower-rank matrix. This reduces the number of parameters in the model and can improve its generalization performance. However, SVD can be computationally expensive and may not be suitable for all architectures.

	 \section{A brief overview of RMT}
	 \subsection{Marchenko–Pastur distribution}

  One of the most important results in RMT is based on the Marchenko-Pastur (MP) distribution. The MP distribution is a probability distribution that arises in the study of random matrices. It is a fundamental result in random matrix theory and has found applications in various fields, including signal processing, wireless communications, and machine learning, see \cite{vershynin2018high,ge2021large,serdobolskii2000multivariate,couillet2011random}. The distribution is used to describe the limiting spectral density of large, random matrices. It provides information about the asymptotic distribution of eigenvalues in a random matrix and predicts the behavior of random matrices under different conditions. The MP distribution is also used in principal component analysis (PCA) and other dimension reduction techniques, see \cite{abdi2010principal,bro2014principal,ringner2008principal}.

 First, we define the empirical spectral distribution (ESD) of a $N \times M$ matrix $G$ by:
  \begin{defn} \label{ESD Def}
    \begin{equation}   \mu_{G_M}=\frac{1}{M}\sum_{i=1}^M \delta_{\sigma^*_i},    \end{equation}
    with $\sigma^*_i$ the $i$th singular value of $G$ and $\delta$ the Dirac measure.
     \end{defn}
	 \begin{thm}[Marchenko and Pastur (1967) \cite{marchenko1967distribution}]
\label{RMT_MP_theorem}	 
Take $W$ to be a $N\times M$ random matrix  for $M \leq N$ with $W_{i,j}$ independent identically distributed random variables from a distribution with mean $0$ and variance $\sigma^2<\infty$. Take  $X =\frac{1}{N} W^T  W$, then assuming $N \to \infty$ and $\frac{N}{M} \to c$ the ESD of $X$, $\mu_{X_M}$, converges weakly in distribution to the Marchenko-Pastur probability distribution given by: 

\begin{equation}
\label{MP_dis}
 {\frac {1}{2\pi \sigma ^{2}}}{\frac {\sqrt {(\lambda _{+}-x)(x-\lambda _{-})}}{c x}}\,\mathbf {1} _{x\in [\lambda _{-},\lambda _{+}]}\,dx
\end{equation}

with

\begin{equation}
\label{lambda}
 \lambda _{\pm }=\sigma ^{2}(1\pm {\sqrt {c }})^{2}.
\end{equation} 

\end{thm}

The theorem states that as the dimensions of the random matrix grow, the distribution of its eigenvalues converges to the Marchenko-Pastur distribution. The MP distribution is a deterministic distribution that depends on two parameters: the variance of the random variables in the original matrix, $\sigma^2$, and the ratio of the number of columns to the number of rows, $c$.

The theorem provides important insights into the structure of large random matrices and is widely used in deep learning to study overfitting and the generalization of deep neural networks. For example, it can be used to determine the conditions under which overfitting occurs and to develop methods for controlling it, see \cite{meng2023impact}. 

\subsection{Spiked Model}

 A spiked model is a model of a matrix which has some deterministic structure and some random structure. One such model is known as the information plus noise model. In this model an $N \times M$ matrix $W$ is given by:  \begin{equation}
      W=S+R,  
    \end{equation} 
    with $R$ random and $S$ a deterministic matrix. One way of studying such a model is by looking at the ESD of the $M \times M$ matrix   $X=\frac{1}{N}W^T  W$.
  
Under certain conditions, the large eigenvalues of $X$ (i.e. the eigenvalues bigger than  $\lambda_+$ given in \eqref{lambda})  correspond to  the singular  values of $S$ with some deterministic perturbation and are called \emph{spikes}, see \cite{benaych2011eigenvalues}.  They "bleed out" of the MP distribution to the right.

Thus, the eigenvalues of the spiked model reflect both the deterministic and random structure of the matrix. The study of spiked models is relevant in many applications such as PCA, low rank matrix completion, and portfolio optimization. In essence, the spiked model gives us a way to understand how the randomness and structure interact in a matrix. The spikes in the ESD correspond to the deterministic structure of the matrix $S$, and the behavior of the spikes can help us understand how much structure the matrix has compared to randomness.

This makes the spiked model a useful tool for various problems in machine learning and signal processing, such as PCA, blind source separation, and low-rank matrix recovery. The idea is that the spikes in the ESD represent the important directions or structures in the data, and by studying these spikes one can gain insights into the underlying data.

In practice, the spiked model is also often used to study the robustness of algorithms that use SVD or PCA as a preprocessing step. The idea is to see how the eigenvalue spikes behave under different noise levels and how well the algorithms can still recover the structure in the data. For example, if the spikes are robust to noise, it means that the structure in the data is well-defined and easily recoverable, while if the spikes are easily washed out by noise, it means that the structure is not well-defined and harder to recover.

\begin{ex}\label{deterministic_random}
  In this example, we create a random $N\times N$ matrix $R$  with components taken from i.i.ds using the normal distribution of zero mean and unit variance ($\sigma^2=1$).  We take $S$ to be a  $N\times N$ deterministic matrix  with components given by
  
  \begin{equation}
  S[i,j]=\tan(\frac{\pi}{2}+\frac{1}{j+1})+\cos(i)\cdot\log(i+j+1)+\sin(j)\cdot\cos(\frac{i}{j}),
  \end{equation}
      $W=R+S$ and $X = \frac{1}{N}W^T W$. The BEMA algorithm is used to find the $\lambda_+$ of the ESD of $X$, as described in Subsection \ref{finding_lambda}. $R$ is a random matrix satisfying the conditions of Theorem \ref{RMT_MP_theorem}, and so the ESD of $\frac{1}{N}R^TR$ converges to the Marchenko-Pastur distribution as $N \to \infty$ and has a $\lambda_+$ that determines the rightmost edge of its compact support. We can imagine a situation in which $R$ is not directly known, and the goal is to find an estimator of $\lambda_+$ from the ESD of $X$. See Fig. \ref{fig:ESD_X&MP} for the result of the ESD of $X$ with the Marchenko-Pastur distribution that best fits the ESD shown in red. 
 \begin{figure}[h!]
		\includegraphics[width=.8\textwidth]{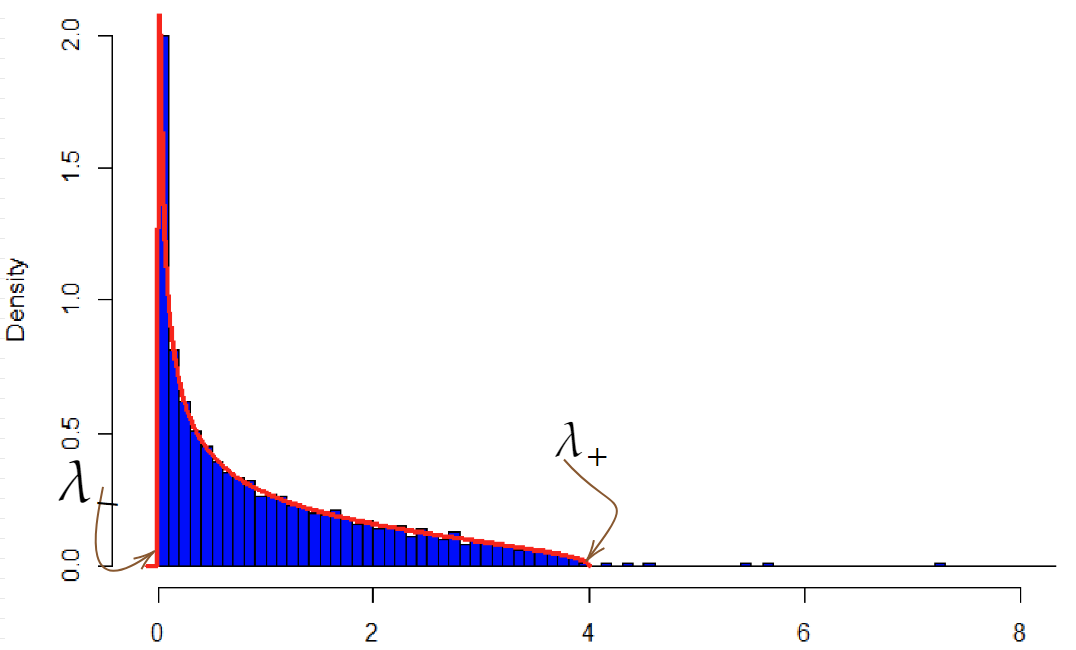}
			
		\caption{In blue we hae the ESD of X, in red the Marchenko-Pastur distribution which best fits the ESD based on the BEMA algorithm.}
		\label{fig:ESD_X&MP}
\end{figure}

 The bulk of the eigenvalues are well-fit by the MP distribution, but some eigenvalues bleed out to the right of $\lambda_+$. These eigenvalues correspond to the singular values of $S$. The direct calculation of the $\lambda_+$ of the MP distribution corresponding to $\frac{1}{N}R^TR$ gives $\lambda_+=\sigma^2\cdot(1+1)^2=4$, and the $\lambda_+$ obtained to fit the bulk of the ESD of $X$ and the $\lambda_+$ of $\frac{1}{N}R^TR$ are approximately the same.

\end{ex}

\subsection{BEMA algorithm for finding $\lambda_+$}
\label{finding_lambda}

The following is the BEMA algorithm for finding best fit $\lambda_+$ of $\frac{1}{N}R^TR$ based on the ESD of $X$. It is used in the analysis of matrices with the information plus noise structure, where one wants to determine the rightmost edge of the compact support of the MP distribution. The BEMA algorithm is computationally efficient and has been shown to provide accurate results for matrices with the information plus noise structure.  The algorithm can be found in \cite{ke2021estimation}. Here we present a simplified version of it for $R$ a $N \times N$ matrix: 

\begin{enumerate}
    \item Choose parameters $\alpha \in (0, 1/2), \beta \in (0,1)$.
    
    \item For each $\alpha N \leq k \leq (1-\alpha)N$, obtain $q_k$, the $(k/N)$ upper-quantile of the MP distribution with $\sigma^2=1$ and $c = 1$.\\
    
    Each $q_k$ is a solution to $\int\limits_{0}^{q_k} \frac{1}{2\pi}\frac{\sqrt{(4-\lambda)\lambda}}{\lambda} = k/N$. 
    \item Compute $\hat{\sigma}^2 = \frac{\sum_{\alpha N \leq k \leq (1-\alpha)N}q_k \lambda_k}{\sum_{\alpha N \leq k \leq (1-\alpha)N}q_k^2}$, where $\lambda_k$ is the $k^{th}$ smallest eigenvalue of $X$.
    
    \item Obtain $t_{1-\beta}$, the $(1-\beta)$ quantile of Tracy-Widom distribution.
    
    \item Return $\lambda_+ = \hat{\sigma}^2[4+2^{4/3}t_{1-\beta}\cdot N^{-2/3}]$.
    
\end{enumerate}
    
    \begin{remark}
   
The algorithm depends on parameters $\alpha \in (0,1/2), \beta \in (0,1)$. We show this by varying $\alpha$ and $\beta$ for the case found in Example \ref{deterministic_random}. See Fig. \ref{dep_alpha} and \ref{dep_beta}. The red  line is $\lambda_+ = 4 $, which is the correct $\lambda_+$ of $\frac{1}{N}R^TR$. In this example, while dependence on $\alpha$ is insignificant for sufficiently large values, dependence on $\beta$ allows us to control the confidence that the eigenvalues of the random matrix $R$ will be smaller than the estimator for $\lambda_+$ of the MP distribution.

\begin{figure}
     \begin{subfigure}{0.4\textwidth}
  \includegraphics[width=\textwidth]{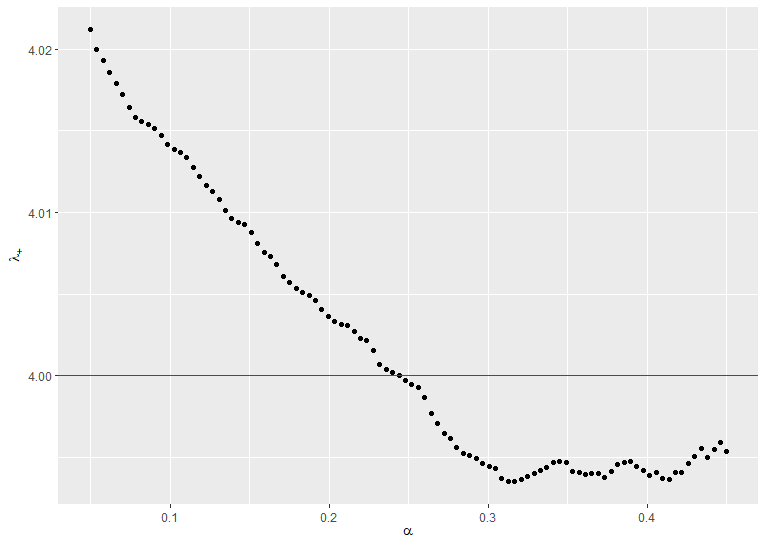}
    \caption{Dependence of algorithm the choice of $\alpha$, $\beta = 0.5$. In this example the rank of the deterministic matrix $S$ is fairly low.}
    \label{dep_alpha}
     \end{subfigure}
     \hfill
     \begin{subfigure}{0.4\textwidth}
  \includegraphics[width=\textwidth]{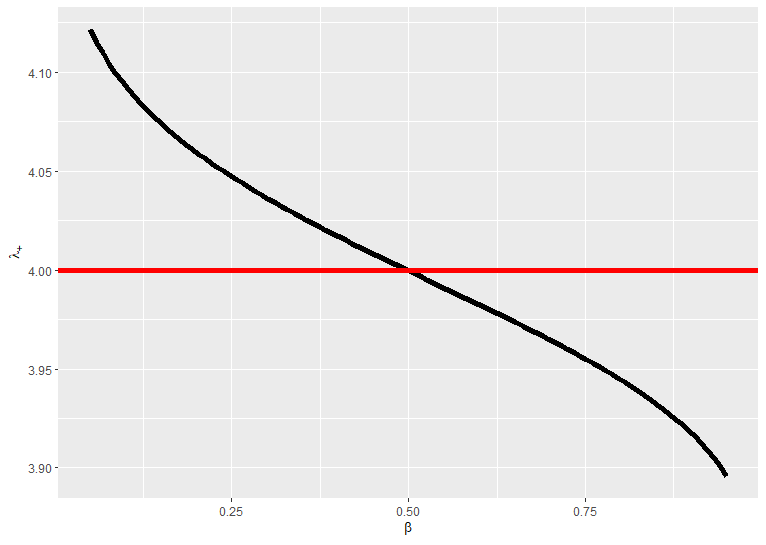}
    \caption{Dependence of algorithm on the choice of $\beta$, $\alpha = 0.25$.}
    \label{dep_beta}
     \end{subfigure}
     \end{figure}

    \end{remark}

\section{ Random matrix theory in deep learning.}

As stated in Section \ref{into}, a DNN is a composition of affine functions $M_l$ and non-linear activation functions. The affine functions $M_l$ can be thought of as a $N \times M$ matrix $W_l$ of parameters and a bias vector $\beta_l$. In this work, we only focus on the matrix $W_l$ of parameters. It has been shown that  $W_l$ can be studied using the spiked model approach in random matrices, with the ESD of $X=\frac{1}{N}W^T  W$ having some eigenvalues which are bigger than $\lambda_+$ and some eigenvalues which are smaller than $\lambda_+$, see \cite{martin2021implicit,staats2022boundary}. 

More specifically, take $X_l(t)=\frac{1}{N}W_l(t)^T W_l(t)$, with $W_l(t)$ a $N \times M$ weight of $l$th layer matrix at time $t$ of DNN training. Assuming that $W_l(t)=R_l(t)+S_l(t)$, with $R_l(t)$ random and $S_l(t)$ a deterministic matrix, we use RMT to study this spiked model of $W_l(t)$. One can assume that during training we go from $W_l(0)=R_l$ (i.e. $W_l$ is random) to $W_l(t_\text{final})=R_l(t_\text{final})+S_l(t_\text{final})$, with $||S_l(t_\text{final})|| \neq 0$ and $t_\text{final}$ the final training time.  
Meaning that as $t \to t_\text{final}$, $||S_l(t)||$ grows and so $W_l(t)$ becomes less random.

We will use the BEMA algorithm to estimate the value of $\lambda_+$ from the ESD of $X_l(t)$. As the training of DNN progresses, the eigenvalues of $X_l(t)$ can for the most part be expected to fit the MP distribution. However, some of the eigenvalues may bleed out of the bulk of the MP distribution and correspond to the singular values of $S_l(t)$. The BEMA algorithm aims to find the rightmost edge of the MP distribution, which determines the value of $\lambda_+$. This is important as $\lambda_+$ can provide insight into the behavior of the DNN during training and its ability to generalize to unseen data.

The BEMA algorithm can be used in conjunction with the SVD to determine which singular values of the weight matrices $W_l$ of the DNN should be removed during training. The SVD decomposes the weight matrix into its singular values and singular vectors, which can then be analyzed using RMT to determine their distribution. By using the BEMA algorithm, one can identify the eigenvalues that correspond to the singular values of $S_l$ and distinguish them from the eigenvalues that correspond to the singular values of $R_l$. These eigenvalues that correspond to $R_l$ can then be removed, allowing for a more effective and efficient training process for the DNN.

\subsection{Singular value decomposition in deep learning}

Take $A$ to be a $N \times M$  matrix. A singular value decomposition of $A$ is a factorization
$A = U\Sigma V^T$ where:

\begin{itemize}
    \item $U$ is an $N \times N$ orthogonal matrix.
    \item $V$ is an $M \times M$ orthogonal matrix.
     \item $\Sigma$ is an $N \times M$ matrix whose $i$
th diagonal entry equals the $i$
th singular
value $\sigma_i$ and all other entries of $\Sigma$ are zero. 
\end{itemize}

 For  $\lambda_i$ the eigenvalues of a matrix $X=W^T  W$ we have that $\sigma_i=\sqrt{\lambda_i}$ are the singular values of $W$. Thus, singular values are related to eigenvalues of the symmetrization of a matrix $W$. 
 
 For $W_l$ of a DNN, it has been shown that removing the small singular values of $W_l$, via its SVD, during the training of a DNN can reduce the number of parameters of the DNN while increasing accuracy, see \cite{yang2020learning,xue2013restructuring,cai2014fast,anhao2016svd}. In the reminder of this work, we show how using RMT can help determine which are the singular values to remove from a DNN layer so as to not decrease the accuracy of the DNN. 

 Specifically, the BEMA algorithm can be used in combination with the SVD of $W_l$ to determine which singular values to remove from the DNN during training. To do this, one first computes the SVD of $W_l$ and then calculates the eigenvalues of the symmetrized matrix $X_l=\frac{1}{N}W_l^TW_l$. The eigenvalues obtained from the symmetrization can then be related to the singular values of $W_l$ through $N\lambda_i=\sigma_i^2$. Using the BEMA algorithm to estimate the value of $\lambda_+$, one can then determine a threshold for the singular values of $W_l$. The singular values smaller than the threshold can be removed without affecting the accuracy of the DNN, as they are likely to be less important for the DNN's performance. This can be done iteratively during the training of the DNN, as the threshold can be updated as the training progresses. By using the BEMA algorithm to determine the threshold, one can effectively reduce the number of parameters in the DNN while still preserving its accuracy. 
 
 \subsection{Removing singular values without decreasing accuracy}
 \label{stable_acc}
 
 In this subsection, we show how SVD can be used to 'cut' out the random parts of $W_l$ without decreasing accuracy. This might allow us to drastically reduce the number of parameters in the DNN resulting in possibly faster training and less overfitting.

 \begin{enumerate}
     \item Obtain a weight matrix $W_l$ of a trained DNN.

\item Perform SVD of $W_l$:  $W_l = U \Sigma V^T$.

\item Compute the eigenvalues $\lambda_i$ of the square matrix $ \frac{1}{N} W^T_l W_l$.

\item Use the BEMA algorithm from Subsection \ref{finding_lambda} to find the best fit MP distribution for the ESD of $X=\frac{1}{N}W_l^TW_l$ and its corresponding $\lambda_+$.

\item Determine if the ESD of $X$ fits the MP distribution. An algorithm for this is given in Subsection \ref{EDS_fit}.   

\item Determine the number of eigenvalues that fall inside of the MP distribution.

\item Remove the smallest singular values in the SVD decomposition equal to the number of eigenvalues determined in step 6.

\item Replace the singular values with zeros to obtain a new diagonal matrix $\Sigma'$.

\item Obtain two layers $W'_l=U\sqrt{\Sigma'}$ and $W''_l=\sqrt{\Sigma'}V^T$ from the SVD decomposition.

\item Replace the original weight matrix $W_l$ with the two new layers $W'_l$ and $W''_l$, with no activation function between them.

\end{enumerate}

The algorithm presented here is a method for reducing the number of parameters in a DNN while maintaining its accuracy. The algorithm uses the technique of SVD to factorize the weight matrix of a trained DNN, $W_l$, into three matrices: $U$, $\Sigma$, and $V^T$. The diagonal matrix $\Sigma$ contains the singular values of $W_l$, which are related to the eigenvalues of the square matrix $W^T_l W_l$.

The algorithm then applies the BEMA algorithm to determine which singular values are likely to correspond to the random parts of the weight matrix. This is done by fitting the ESD of $\frac{1}{N}W_l^TW_l$ to a MP distribution and obtaining its corresponding $\lambda_+$. If a certain number of the eigenvalues "bleed" out of the MP distribution, then it is assumed that these correspond to the non-random parts of the weight matrix.

Finally, the algorithm sets the smallest singular values that are considered to be the random parts of the weight matrix to 0 and replaces the original weight matrix with two new layers, $W'_l=U\sqrt{\Sigma'}$ and $W''_l=\sqrt{\Sigma'}V^T$, obtained from the modified $\Sigma'$. The assumption is that by removing the small singular values, only the random parts of the DNN are being removed, and thus the accuracy of the DNN should remain the same.

Overall, this algorithm presents a promising way to reduce the number of parameters in a DNN while preserving its accuracy, as it takes advantage of the underlying structure of the weight matrices to determine which parts can be safely removed. However, a more detailed justification of the algorithm's effectiveness will require further research.

 \begin{ex}
 We used the above approach for a DNN trained on MNIST. In this example the DNN has two layers, the first with a $784 \times 1000$ matrix $W_1$ and the second with a $1000 \times 10$ matrix $W_2$. The activation function was ReLU. We trained the DNN for $10$ epocs and achieved a $98$\% accuracy on the test set.

\begin{figure}%
    \centering
    \subfloat[\centering Full Empirical Density]{{\includegraphics[width=6.5cm]{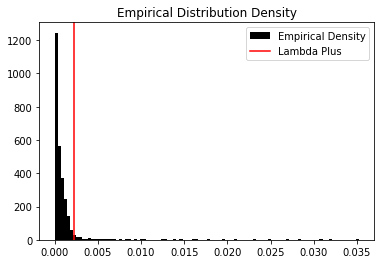} }}%
    \qquad
    \subfloat[\centering Zoomed Density ]{{\includegraphics[width=6.5cm]{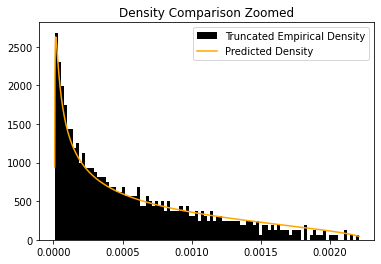} }}
    \caption{The ESD of $X_l$ and its best fit MP distribution}
    \label{MPdis1}%
\end{figure}

       In Fig. \ref{MPdis1}  the  ESD of $X_1=\frac{1}{N}W^T_1W_1$ is shown together with its best fit MP distribution. Most eigenvalues of $X_1$ lie inside the MP distribution. We perform a SVD on $W_1$, in this case $\Sigma$ is a $784 \times 1000$ matrix. Even if we only keep the biggest $20$ $\sigma_i$ of $W_1$ and transform the first layer into two layers $W'_1$ and $W''_1$ the accuracy is still $92$\%. $W_1$ had $784,000$ parameters, while $W'_1$ and $W''_1$ have $15,680+20,000=35,680$ parameters (not including the bias vector parameters). This is a reduction by over $90\%$. In Fig. \ref{pvalue} we show how the accuracy of the DNN depends on the number of singular values which we keep. The red line corresponds to the threshold given by the MP distribution (via $\lambda_+$) for how many of the large singular values should be kept. As the figure shows, this threshold is highly accurate.   
       
       \begin{figure}[h!]
			\includegraphics[width=.5\textwidth]{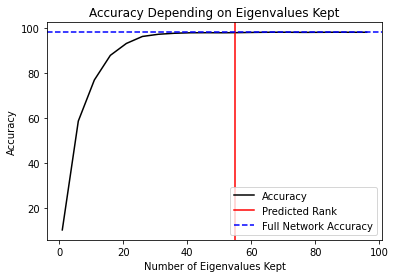}
		\caption{Number of eigenvalues kept is shown on the $x$-axis while the accuracy is shown on the $y$-axis.}
		\label{pvalue}
		\end{figure}

 \end{ex}

 \subsection{Does the ESD of $X$ fit a spiked MP distribution?}
 \label{EDS_fit}
In this subsection, we outline an algorithm to determine whether the ESD of \(X\) is likely to have came from a certain MP distribution (potentially with spiked eigenvalues). The basis of this algorithm is formed by first using the BEMA algorithm to determine the best fitting MP distribution. This best fitting distribution gives a theoretical cumulative distribution function, and we can compute the cumulative empirical spectral distribution associated with \(X\). These two distributions can then be compared and we can reject the claim that \(X\) follows the predicted MP distribution if the two distributions differ by too much. We make these notions precise now, beginning with the definition of an empirical cumulative spectral distribution.
\begin{remark}
    As with any sort of statistical test, we can never prove or assert with certainty that given empirical data actually was generated according to a prescribed distribution. Imagine flipping a fair coin 1000 times, and by chance, the coin comes up heads every time. Any observer who did not know the coin was fair would rightfully conclude the coin was biased with high probability, but of course, we know the coin to be fair. In this way, statistical tests can indicate that data was not generated according to a certain distribution with high probability, but they cannot conclusively state this.
\end{remark}
\begin{defn} \label{CED}
    Suppose \(G\) is a \(N \times M\) matrix and its ESD \(\mu_{G_M}\) is defined as in Definition \ref{ESD Def}. Then the empirical cumulative spectral distribution of \(G\), \(F_G: \R \to \R\), is defined as follows: 
    \begin{equation}
        F_G(a) = \mu_{G_m}((-\infty, a])
    \end{equation}
\end{defn}
As it turns out, the cumulative distribution functions for the MP distribution are known with a closed form. With these formulas, we are now ready to explain our algorithm in full. We fix a tuning parameter \(\gamma \in (0,1)\) which corresponds to the sensitivity of our test.
\begin{enumerate}
    \item Take as input \(X = \frac{1}{N} W^T W\) where \(W\) is an \(N \times M\) matrix, and compute the spectrum of \(X = \{\sigma_1, \dots, \sigma_M\}\).
    \item Compute the empirical cumulative spectral distribution of \(X\), denoted \(F_X\).
    \item Perform the BEMA algorithm with parameters \(\alpha\) and \(\beta\) to determine \(\hat\sigma^2\), the predicted variance of each coordinate of $W$.
    \item Compute \(0 \leq i_{\text{low}} < i_{\text{high}} \leq M\) such that \(i_{\text{low}}\) is the smallest integer with \(\frac{i_{\text{low}}}{M} \geq \alpha\) and similarly \(i_{\text{high}}\) is the largest integer with \(\frac{i_{\text{high}}}{M} \leq 1 - \alpha\).
    \item Define \(F_X'\) to be the theoretical cumulative distribution function for the MP distribution with parameters \(\hat \sigma^2\) and \(\lambda = N/M\).
    \item Compute \(s = \max_{i \in [i_{\text{low}}, i_{\text{high}}]} \left |F_X(i) - F_X'(i) \right |\).
    \item If \(s > \gamma\) we reject the claim that \(X\) follows the given distribution. If \(s \leq \gamma\) we do not reject this claim.
\end{enumerate}
In words, this algorithm computes the max difference between the predicted and empirical cumulative distribution functions by sampling at each point in the empirical distribution. Since this is to be applied for the specific case of testing for spiked MP distributions, we can use this information to improve our test over naively computing the \(L^\infty\) difference between the predicted and empirical distributions.

This improvement comes in the step which computes \(i_{\text{low}}\) and \(i_{\text{high}}\). Since BEMA only uses data in the quantile between \((\alpha, 1 - \alpha)\) to find the best fit, it makes sense to only test for fit in the same range. In context, we would expect a spiked MP distribution to be poorly approximated by its generating MP distribution around the biggest eigenvalues (i.e., the spiked values), and hence it makes sense to only test the bulk values for goodness of fit.

 \subsection{RMT algorithm for training DNNs}
 \label{RMT_DNN_algorithm}
 
The following outlines the steps for implementing a DNN algorithm that helps prevent overfitting.

\begin{enumerate}
    \item Begin by training the DNN for a set number of epochs, denoted as $\ell$.
\item After $\ell$ epochs, perform a singular value decomposition SVD on the layers of the DNN.
Based on the criteria from Subsection \ref{stable_acc}, remove a portion, for example $45\%$, of the small singular values.
\item Split the layer into two new layers, as described in Subsection \ref{stable_acc}.
\item Only proceed with steps (2)-(4) if the new layers have fewer parameters than the original layer and the ESD of $X=\frac{1}{N}W^T_lW_l$ fits the MP distribution as described in Subsection  \ref{EDS_fit}.
\item Continue training the DNN using the new layers. This completes one cycle of training.
\item Determine a new value for $\ell$ to represent the number of epochs between cycles.
\item Repeat steps (2)-(5) every $\ell$ epochs.

\end{enumerate}

It is important to note that in step $2$ we don't remove all of the small singular values (i.e. singular values whose corresponding eigenvalues are inside of the MP distribution).  We found that it is crucial to strike a balance between removing the small singular values and retaining some of them. As mentioned earlier, removing all of the small singular values might lead to underfitting of the DNN, and thus hinder its ability to learn the underlying data patterns. On the other hand, keeping some of the small singular values introduces some randomness in the weight layer matrix $W_l$, which we found to be beneficial for the DNN's performance. Therefore, having an appropriate RMT threshold to determine which singular values to retain and which ones to remove is beneficial when optimizing the DNN's learning ability.

 \subsection{Numerical results on MNIST}

 We performed the algorithm given in Subsection \ref{RMT_DNN_algorithm} on a DNNs trained on MNIST. We deliberately overparameterized the DNNs so that they overfit and trained them to achieve a nearly $100\%$ accuracy on the training set. We used DNNs with the ReLU activation, the cross-entropy loss function function and a step size of $.05$. 

 \begin{remark}
     In our experiments, we seeded the weight matrices of the DNN with uniformly distributed weights drawn from the range $[-1/\sqrt{n},1/\sqrt{n}]$ , where $n$ is the number of inputs to the layer. This seeding method has been shown to work well in our experiments, and is a commonly used technique in deep learning for initializing weight matrices.

 \end{remark}

 \begin{ex}
 In this example, we took a DNN with $4$  layers. The first layer had a size $784 \times 3000$, the second $3000 \times 3000$, the third $3000 \times 500$ and the fourth $500 \times 10$. We denote such a DNN by $[784,3000,3000,500, 10]$. It is important to  note that, in all examples, the final layer of the DNN will have a relatively small size (in this case $500 \times 10$). This is because the ESD of the symmetrization of the final layer normally does not fit the MP distribution and so will never change. We, therefore, avoid putting too many parameters in the layer so as to not overfit on its account.

 We train two DNNs for $90$ epochs. This first, called non-split, is a normal DNN. The second, called split, performs the algorithm given in Subsection \ref{RMT_DNN_algorithm} and removes $45\%$ of the singular values in each layer every $3$ epochs and when all conditions are satisfied. Recall, one of the conditions is that the two new layers, which are formed out of an old layer, must have fewer parameters than the old layer. This ensures that the split DNN has fewer parameters than the non-split one. Another condition is that the ESD of the symmetrization of the layers fits the MP distribution, as described in Subsection \ref{EDS_fit}. We verified numerically that this  ensures the accuracy of the DNN does not decrease when we split a layer into $2$. We required a $0.15$ goodness of fit, (see Subsection \ref{EDS_fit}). 
 
 Fig. \ref{DNN_new_algo_ex_1} shows the accuracy of both DNNs on a test set. We see that the accuracy of the non-split DNN plateaus at around $86$, a sign that it is overfitting, while the accuracy of the split DNN peaks at $97.5$. Finally, the number of parameters of the non-split DNN was $12,863,510$ while the number of parameters of the split DNN was $3,881,124$ (including the parameters of the bias vectors). In fact, the split DNN can be represented by: $$[784, 367, 176, 89, 54, 35, 3000, 143, 293, 627, 1366, 629, 294, 139, 3000, 71, 124, 241, 500, 10].$$     

 \begin{figure}[h!]
			\includegraphics[width=.5\textwidth]{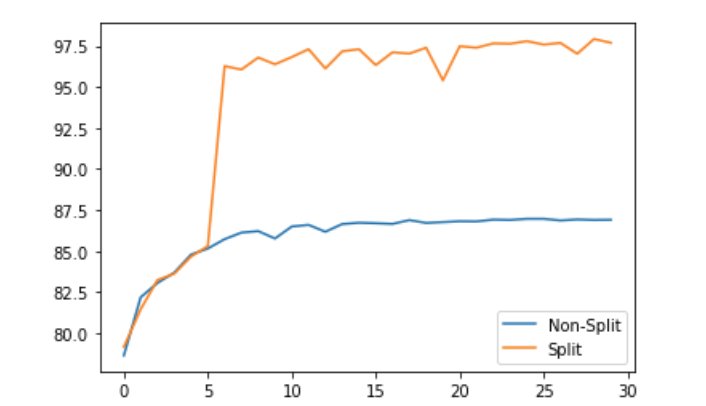}
		\caption{Accuracy of split and non-split algorithms on the test set.}
		\label{DNN_new_algo_ex_1}
		\end{figure}

 \end{ex}

 \begin{ex}
 \label{good_example}
 In this example, we stated with a DNN with $5$  layers. We denote this DNN by $[784,1500,1500, 500, 500, 10]$. We train two DNNs for $300$ epochs. The first, called non-split, is a normal DNN. The second, called split, performs the algorithm given in Subsection \ref{RMT_DNN_algorithm} and removes $45\%$ of the singular values in each layer every $3$ epochs and when all conditions are satisfied. We require that the ESD of the symmetrization of the layers fits the MP distribution, as described in Subsection \ref{EDS_fit}. In this example, we required a $ 0.012$ goodness of fit. In Example \ref{bad_example}, we start with the same original DNN but require a much weaker goodness of fit of $.1$ to illustrated some aspects of this hyper-parameter.   
 
 Fig. \ref{DNN_example_split_1} shows the accuracy of both DNNs on a test set. We see that the accuracy of the non-split DNN plateaus at around $88$, a sign that it is overfitting, while the accuracy of the split DNN peaks at $97$. The number of parameters of the non-split DNN was $4,435,010$ while the number of parameters of the split DNN was $2,554,157$. In fact, the split DNN can be represented by: $$[784, 354, 167, 1500, 309, 677, 309, 1500, 228, 500, 105, 227, 107, 500, 10].$$     

 \begin{figure}[h!]
			\includegraphics[width=.5\textwidth]{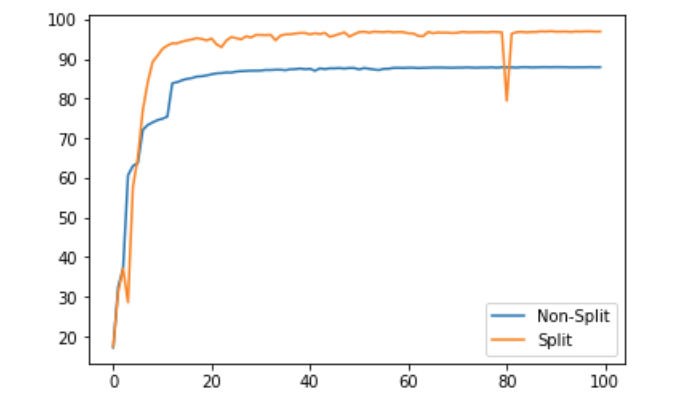}
		\caption{Accuracy of split and non-split algorithms on the test set.}
		\label{DNN_example_split_1}
		\end{figure}

 Finally, in Table \ref{ESD_examples_false} and Table \ref{ESD_examples_true}   we show the ESD of all of the layers of the split DNN after its second cycle, when it looked like $[784, 354, 1500, 677, 1500, 228, 500, 227, 500, 10]$, with the $\lambda_+$ obtained from the BEMA algorithm. Table \ref{ESD_examples_false}, shows the ESD of the weight layer matrices of the split DNN which did not fit the MP distribution and so were not split into two layers. Table \ref{ESD_examples_true} shows the ESD of the weight layer matrices which did fit the MP distribution. Some of them were split into two layers, based on the criteria that the two new layers must have less parameters than the original layers, and some were not split into two new layers.   The new DNN looked like: $$[784, 354, 1500, 309, 677, 309, 1500, 228, 500, 105, 227, 107, 500, 10]$$.

 \begin{longtable}{ccc}
    \caption{ESD of the weight layer matrices of the split DNN which did not fit the MP distribution and so were not split into $2$.} \\
\label{ESD_examples_false}
\includegraphics{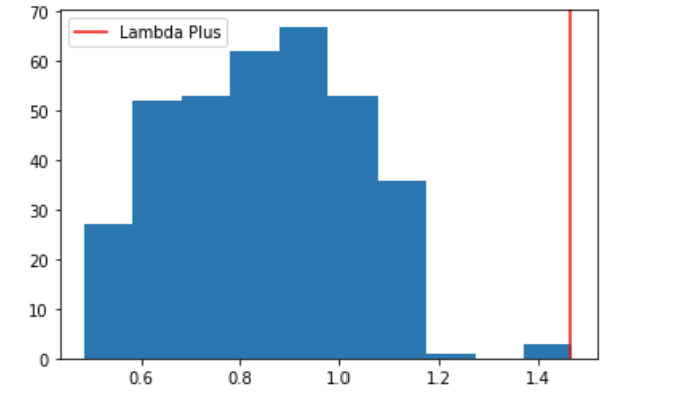}
    & \includegraphics{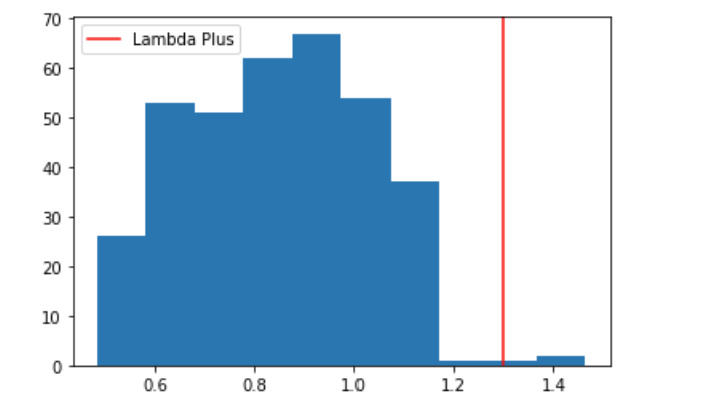} 
    & \includegraphics{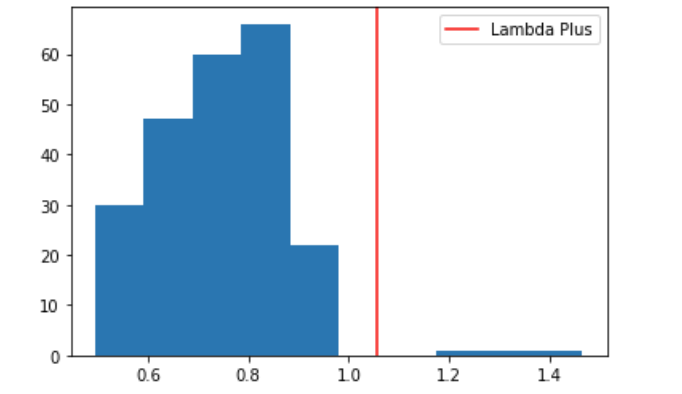} 
    \cr
\includegraphics{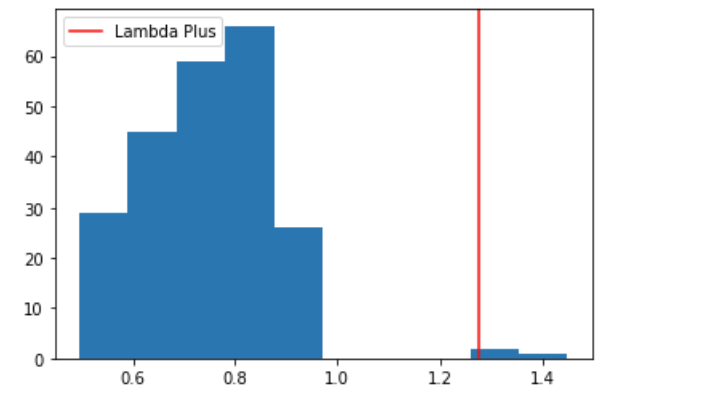} 
&
\includegraphics{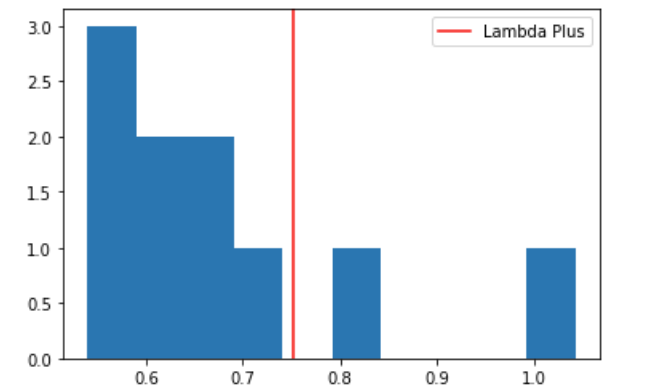}
\end{longtable}

\begin{longtable}{ccc}
    \caption{ESD of the weight layer matrices of the split DNN which did fit the MP distribution. } \\
\label{ESD_examples_true}
\includegraphics{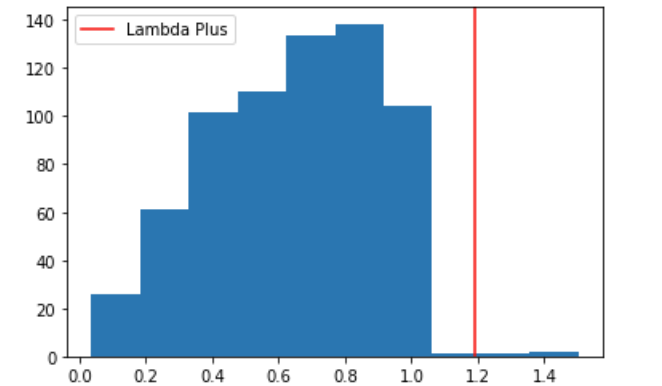}
    & \includegraphics{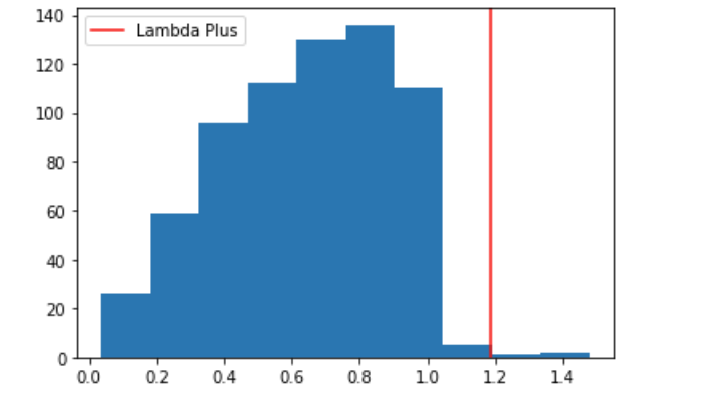} 
     \cr
\includegraphics{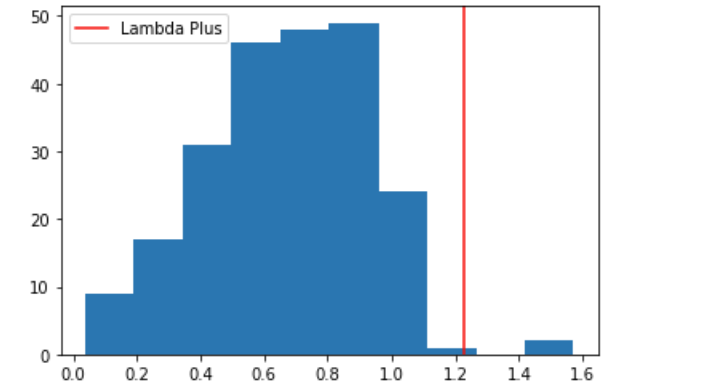} 
   &
\includegraphics{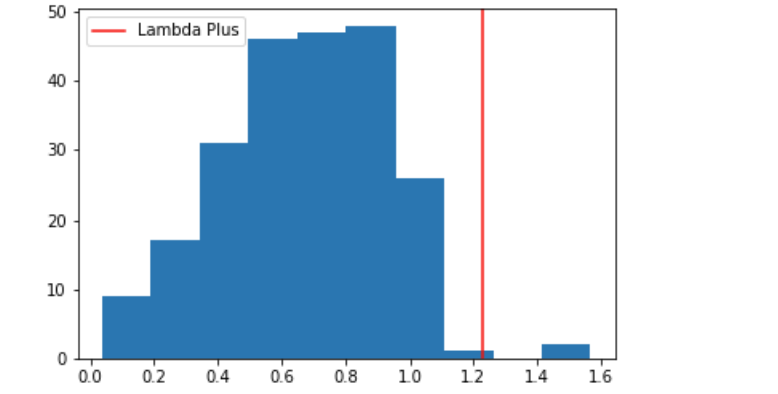} 

\end{longtable}

 \end{ex}

 \begin{ex}
 \label{bad_example}
In this example, we took the same DNN with $5$  layers as in Example \ref{good_example}. We denote this DNN by $[784,1500,1500, 500, 500, 10]$. We train two DNNs for $300$ epochs. The first, called non-split, is a normal DNN. The second, called split, performs the algorithm given in subsection \ref{RMT_DNN_algorithm} and removes $45\%$ of the singular values in each layer every $3$ epochs and when all conditions are satisfied. We require that the ESD of the symmetrization of the layers fits the MP distribution, as described in subsection \ref{EDS_fit}. This ensures that the accuracy of the DNN does not decrease when we split a layer into $2$. In this example, we required a $ 0.1$ goodness of fit. 
 
 Fig. \ref{DNN_example_split_1} shows the accuracy of both DNNs on a test set. We see that the accuracy of the non-split DNN plateaus at around $88$, a sign that it is overfitting, while the accuracy of the split DNN peaks at $95$ and then falls to $20$. This might be a sign that the split DNN first found a good number of parameters needed to learn the test set but then because the goodness of fit parameter was not small enough, some important information was lost during some of the splits. The number of parameters of the non-split DNN was $4,435,010$ and the number of parameters of the split DNN was $1,943,563$. The final split DNN can be represented by: $$ [784, 354, 164, 1500, 144, 309, 677, 309, 144, 1500, 107, 228, 500, 51, 107, 227, 107, 53, 500, 10].$$     

 \begin{figure}[h!]
			\includegraphics[width=.5\textwidth]{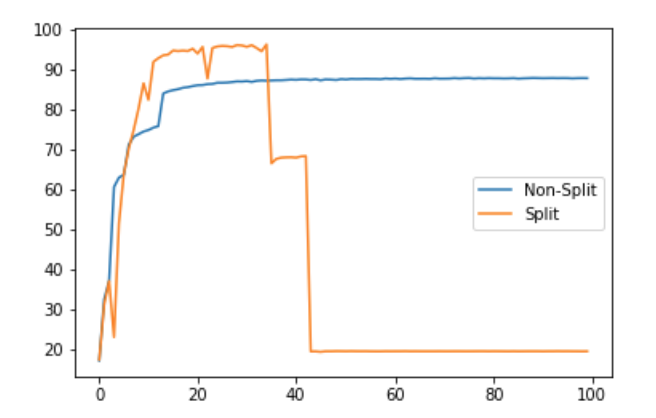}
		\caption{Accuracy of split and non-split algorithms on the test set.}
		\label{DNN_example_split_5}
		\end{figure}     
 \end{ex}

 \begin{ex}

 In this example, we started with a DNN with $6$ layers. We denote the  DNN by $ [784,2500,2500,2500,2500, 500, 10]$. We train two DNNs for $30$ epochs, the non-split and split versions. The first is a normal DNN. The second performs the algorithm given in Subsection \ref{RMT_DNN_algorithm} and removes $45\%$ of the singular values in each layer every $3$ epochs and when all conditions are satisfied. We required a $0.05$ goodness of fit on a layer for it to be split. 
 
 Fig. \ref{DNN_new_algo_ex_2} shows the accuracy of both DNNs on a test set. We see that the accuracy of the non-split DNN plateaus at around $86.4$ (with $51,887/60,000$ objects in the test set classified correctly), a sign that it is overfitting, while the accuracy of the split DNN peaks at $98.6$ (with $59,172/60,000$ objects in the test set classified correctly). Finally, the number of parameters of the non-split DNN was $21,975,510$ while the number of parameters of the split DNN was $13,207,896$. In fact, the split DNN can be represented by: $$[784, 362, 2500, 518, 1134, 518, 2500, 518, 1134, 518, 2500, 518, 1134, 518, 2500, 234, 500, 10].$$   

 \begin{figure}[h!]
			\includegraphics[width=.5\textwidth]{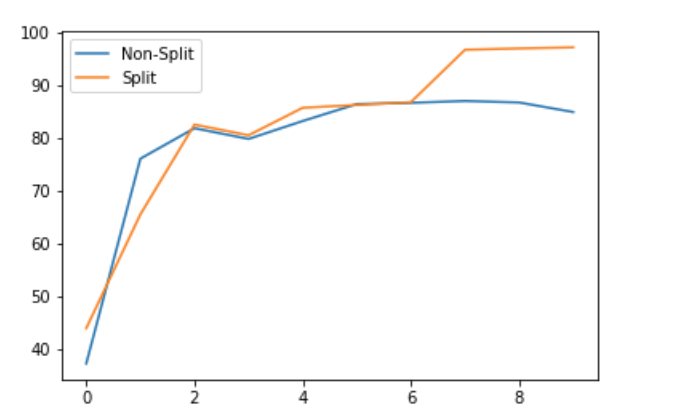}
		\caption{Accuracy of split and non-split algorithms on the test set.}
		\label{DNN_new_algo_ex_2}
		\end{figure}
     
 \end{ex}

 \begin{ex}

 Next we present an example in which the split DNN and non-split DNN both overfit. However, the split DNN still performs better.

 In this example, we took a DNN with $4$ layers. We denote the  DNN by $ [784,3000, 3000, 3000, 10]$. We again train two DNNs for $90$ epochs, the non-split and split versions. For the split DNN, we remove $45\%$ of the singular values in each layer every $3$ epoch and when all conditions are satisfied. We required a $0.01$ goodness of fit (see Subsection \ref{EDS_fit}) on a layer for it to be split. 
 
 Fig. \ref{DNN_new_algo_ex_3} shows the accuracy of both DNNs on a test set. We see that the accuracy of the non-split DNN plateaus at around $78$ while the accuracy of the non-split DNN peaks at $88$. Finally, the number of parameters of the non-split DNN was $20,391,010$ while the number of parameters of the split DNN was $17,716,113$. This reduction in parameters is not as much as we would like, which is probably why we are still overfitting. The split DNN can be represented by: $$[784, 361, 3000, 1359, 3000, 1359, 3000, 10].$$ 

 \begin{figure}[h!]
			\includegraphics[width=.5\textwidth]{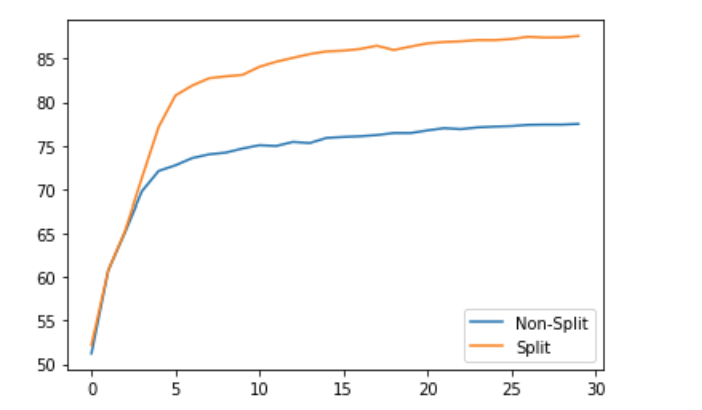}
		\caption{Accuracy of split and non-split algorithms on the test set.}
		\label{DNN_new_algo_ex_3}
		\end{figure}
     
 \end{ex}

 \section{Future Work}

 In this work, we have shown the potential of using RMT for reducing overfitting and improving the accuracy of DNNs. Our experiments on simple DNN models trained on MNIST demonstrate the effectiveness of RMT techniques for regularization, but there are several avenues for further research in this area.

One area of future work is exploring the use of different distributions for seeding the random weight matrices in DNNs. Another area of future work is to develop a better theoretical understanding of the relationship between RMT and overfitting in DNNs. While our experiments demonstrate the effectiveness of RMT for reducing overfitting and improving accuracy, a deeper theoretical understanding of this relationship can help to develop more efficient and effective regularization methods.

Furthermore, applying RMT techniques to different architectures and datasets is another potential avenue for future research. We plan to investigate the scalability of these techniques to larger datasets and more complex models. For example, incorporating RMT-based regularization techniques into other neural network architectures, such as convolutional neural networks or recurrent neural networks, could be an interesting area of investigation. Additionally, exploring the use of RMT in transfer learning scenarios, where a pre-trained network is fine-tuned for a new task, may also prove to be useful.

We also aim to further refine this RMT-based algorithm not only for reducing the complexity of the DNN but also for developing  regularization methods to provide even better performance. This may involve using RMT to determine which singular values should be regularized to obtain better-performing DNNs.

Another area of interest is investigating the impact of RMT on the interpretability of DNNs. As deep learning models become increasingly complex and difficult to interpret, there is a growing need for regularization techniques that can help to promote interpretability. It is possible that RMT-based regularization methods could be used to encourage sparsity or other desirable properties in the learned representations of a neural network, which could in turn aid in interpretability.

In conclusion, the use of RMT techniques in deep learning shows great promise for improving the performance and reliability of DNNs. While there are still many areas for further research, we believe that this approach has the potential to become a key tool for regularization and interpretability in deep learning.



	\bibliographystyle{alpha}

\bibliography{DNNstability_biblio} 

\newcommand{\etalchar}[1]{$^{#1}$}
\begin{thebibliography}{BHMM19}

\bibitem[APJY16]{anhao2016svd}
Xing Anhao, Zhang Pengyuan, Pan Jielin, and Yan Yonghong.
\newblock {S}{V}{D}-based {D}{N}{N} pruning and retraining.
\newblock {\em Journal of Tsinghua University (Science and Technology)},
  56(7):772--776, 2016.

\bibitem[ASS20]{advani2020high}
Madhu~S Advani, Andrew~M Saxe, and Haim Sompolinsky.
\newblock High-dimensional dynamics of generalization error in neural networks.
\newblock {\em Neural Networks}, 132:428--446, 2020.

\bibitem[AW10]{abdi2010principal}
Herv{\'e} Abdi and Lynne~J Williams.
\newblock Principal component analysis.
\newblock {\em Wiley interdisciplinary reviews: computational statistics},
  2(4):433--459, 2010.

\bibitem[BGN11]{benaych2011eigenvalues}
Florent Benaych-Georges and Raj~Rao Nadakuditi.
\newblock The eigenvalues and eigenvectors of finite, low rank perturbations of
  large random matrices.
\newblock {\em Advances in Mathematics}, 227(1):494--521, 2011.

\bibitem[BHMM19]{belkin2019reconciling}
Mikhail Belkin, Daniel Hsu, Siyuan Ma, and Soumik Mandal.
\newblock Reconciling modern machine-learning practice and the classical
  bias--variance trade-off.
\newblock {\em Proceedings of the National Academy of Sciences},
  116(32):15849--15854, 2019.

\bibitem[BS14]{bro2014principal}
Rasmus Bro and Age~K Smilde.
\newblock Principal component analysis.
\newblock {\em Analytical methods}, 6(9):2812--2831, 2014.

\bibitem[CD11]{couillet2011random}
Romain Couillet and Merouane Debbah.
\newblock {\em Random matrix methods for wireless communications}.
\newblock Cambridge University Press, 2011.

\bibitem[CKXS14]{cai2014fast}
Chenghao Cai, Dengfeng Ke, Yanyan Xu, and Kaile Su.
\newblock Fast learning of deep neural networks via singular value
  decomposition.
\newblock In {\em Pacific Rim International Conference on Artificial
  Intelligence}, pages 820--826. Springer, 2014.

\bibitem[CMR21]{cohen2021learning}
Omry Cohen, Or~Malka, and Zohar Ringel.
\newblock Learning curves for overparametrized deep neural networks: A field
  theory perspective.
\newblock {\em Physical Review Research}, 3(2):023034, 2021.

\bibitem[GLBP21]{ge2021large}
Jungang Ge, Ying-Chang Liang, Zhidong Bai, and Guangming Pan.
\newblock Large-dimensional random matrix theory and its applications in deep
  learning and wireless communications.
\newblock {\em Random Matrices: Theory and Applications}, 10(04):2230001, 2021.

\bibitem[HDY{\etalchar{+}}12]{hinton2012deep}
Geoffrey Hinton, Li~Deng, Dong Yu, George~E Dahl, Abdel-rahman Mohamed, Navdeep
  Jaitly, Andrew Senior, Vincent Vanhoucke, Patrick Nguyen, Tara~N Sainath,
  et~al.
\newblock Deep neural networks for acoustic modeling in speech recognition: The
  shared views of four research groups.
\newblock {\em IEEE Signal processing magazine}, 29(6):82--97, 2012.

\bibitem[KKB17]{kawaguchi2017generalization}
Kenji Kawaguchi, Leslie~Pack Kaelbling, and Yoshua Bengio.
\newblock Generalization in deep learning.
\newblock {\em arXiv preprint arXiv:1710.05468}, 2017.

\bibitem[KML21]{ke2021estimation}
Zheng~Tracy Ke, Yucong Ma, and Xihong Lin.
\newblock Estimation of the number of spiked eigenvalues in a covariance matrix
  by bulk eigenvalue matching analysis.
\newblock {\em Journal of the American Statistical Association}, pages 1--19,
  2021.

\bibitem[KSH17]{krizhevsky2017imagenet}
Alex Krizhevsky, Ilya Sutskever, and Geoffrey~E Hinton.
\newblock Imagenet classification with deep convolutional neural networks.
\newblock {\em Communications of the ACM}, 60(6):84--90, 2017.

\bibitem[LBD{\etalchar{+}}89]{LBD}
Yann LeCun, Bernhard Boser, John Denker, Donnie Henderson, Richard Howard,
  Wayne Hubbard, and Lawrence Jackel.
\newblock Handwritten digit recognition with a back-propagation network.
\newblock {\em Advances in neural information processing systems}, 2, 1989.

\bibitem[MBB18]{ma2018power}
Siyuan Ma, Raef Bassily, and Mikhail Belkin.
\newblock The power of interpolation: Understanding the effectiveness of sgd in
  modern over-parametrized learning.
\newblock In {\em International Conference on Machine Learning}, pages
  3325--3334. PMLR, 2018.

\bibitem[MM21]{martin2021implicit}
Charles~H Martin and Michael~W Mahoney.
\newblock Implicit self-regularization in deep neural networks: Evidence from
  random matrix theory and implications for learning.
\newblock {\em The Journal of Machine Learning Research}, 22(1):7479--7551,
  2021.

\bibitem[MP67]{marchenko1967distribution}
Vladimir~Alexandrovich Marchenko and Leonid~Andreevich Pastur.
\newblock Distribution of eigenvalues for some sets of random matrices.
\newblock {\em Matematicheskii Sbornik}, 114(4):507--536, 1967.

\bibitem[MPM21]{martin2021predicting}
Charles~H Martin, Tongsu Peng, and Michael~W Mahoney.
\newblock Predicting trends in the quality of state-of-the-art neural networks
  without access to training or testing data.
\newblock {\em Nature Communications}, 12(1):4122, 2021.

\bibitem[MY23]{meng2023impact}
Xuran Meng and Jianfeng Yao.
\newblock Impact of classification difficulty on the weight matrices spectra in
  deep learning and application to early-stopping.
\newblock {\em Journal of Machine Learning Research}, 24:1--40, 2023.

\bibitem[NKB{\etalchar{+}}21]{nakkiran2021deep}
Preetum Nakkiran, Gal Kaplun, Yamini Bansal, Tristan Yang, Boaz Barak, and Ilya
  Sutskever.
\newblock Deep double descent: Where bigger models and more data hurt.
\newblock {\em Journal of Statistical Mechanics: Theory and Experiment},
  2021(12):124003, 2021.

\bibitem[Pre12]{prechelt2012early}
Lutz Prechelt.
\newblock Early stopping—but when?
\newblock {\em Neural networks: tricks of the trade: second edition}, pages
  53--67, 2012.

\bibitem[Rin08]{ringner2008principal}
Markus Ringn{\'e}r.
\newblock What is principal component analysis?
\newblock {\em Nature biotechnology}, 26(3):303--304, 2008.

\bibitem[Ser00]{serdobolskii2000multivariate}
Vadim~Ivanovich Serdobolskii.
\newblock {\em Multivariate statistical analysis: A high-dimensional approach},
  volume~41.
\newblock Springer Science \& Business Media, 2000.

\bibitem[SHK{\etalchar{+}}14]{srivastava2014dropout}
Nitish Srivastava, Geoffrey Hinton, Alex Krizhevsky, Ilya Sutskever, and Ruslan
  Salakhutdinov.
\newblock Dropout: a simple way to prevent neural networks from overfitting.
\newblock {\em The journal of machine learning research}, 15(1):1929--1958,
  2014.

\bibitem[STR22]{staats2022boundary}
Max Staats, Matthias Thamm, and Bernd Rosenow.
\newblock Boundary between noise and information applied to filtering neural
  network weight matrices.
\newblock {\em arXiv preprint arXiv:2206.03927}, 2022.

\bibitem[SVL14]{sutskever2014sequence}
Ilya Sutskever, Oriol Vinyals, and Quoc~V Le.
\newblock Sequence to sequence learning with neural networks.
\newblock {\em Advances in neural information processing systems}, 27, 2014.

\bibitem[Ver18]{vershynin2018high}
Roman Vershynin.
\newblock High-dimensional probability by roman vershynin, 2018.

\bibitem[XLG13]{xue2013restructuring}
Jian Xue, Jinyu Li, and Yifan Gong.
\newblock Restructuring of deep neural network acoustic models with singular
  value decomposition.
\newblock In {\em Interspeech}, pages 2365--2369, 2013.

\bibitem[XLZ{\etalchar{+}}19]{xu2019trained}
Yuhui Xu, Yuxi Li, Shuai Zhang, Wei Wen, Botao Wang, Wenrui Dai, Yingyong Qi,
  Yiran Chen, Weiyao Lin, and Hongkai Xiong.
\newblock Trained rank pruning for efficient deep neural networks.
\newblock In {\em 2019 Fifth Workshop on Energy Efficient Machine Learning and
  Cognitive Computing-NeurIPS Edition (EMC2-NIPS)}, pages 14--17. IEEE, 2019.

\bibitem[YTW{\etalchar{+}}20]{yang2020learning}
Huanrui Yang, Minxue Tang, Wei Wen, Feng Yan, Daniel Hu, Ang Li, Hai Li, and
  Yiran Chen.
\newblock Learning low-rank deep neural networks via singular vector
  orthogonality regularization and singular value sparsification.
\newblock In {\em Proceedings of the IEEE/CVF conference on computer vision and
  pattern recognition workshops}, pages 678--679, 2020.

\end{thebibliography}
	
\end{subsection}

\end{document}